\documentclass[11pt]{article}

\usepackage[final]{acl}

\usepackage{times}
\usepackage{latexsym}

\usepackage[T1]{fontenc}

\usepackage[utf8]{inputenc}

\usepackage[protrusion=false]{microtype}

\usepackage{inconsolata}

\usepackage{graphicx}

\usepackage{tabularx}
\usepackage{multirow}
\usepackage{soul}
\usepackage{graphicx}
\usepackage{booktabs}
\usepackage{rotating}
\usepackage[table]{xcolor}
\usepackage{color, colortbl}
\usepackage{changepage,threeparttable}
\usepackage[htt]{hyphenat}
\usepackage{hyperref}
\usepackage{afterpage}
\usepackage{float}
\usepackage{changepage}
\usepackage{setspace}
\usepackage{amsmath}
\usepackage{enumitem}  

\usepackage{booktabs}

\newcommand{\accuracycell}[1]{%
  \begingroup
  \def\rawvalue{#1}%
  \ifdim \rawvalue pt < 52pt
    \cellcolor[HTML]{FF9966}\rawvalue %
  \else\ifdim \rawvalue pt < 54pt
    \cellcolor[HTML]{FFA070}\rawvalue %
  \else\ifdim \rawvalue pt < 56pt
    \cellcolor[HTML]{FFA77A}\rawvalue %
  \else\ifdim \rawvalue pt < 58pt
    \cellcolor[HTML]{FFAE84}\rawvalue %
  \else\ifdim \rawvalue pt < 60pt
    \cellcolor[HTML]{FFB58E}\rawvalue %
  \else\ifdim \rawvalue pt < 62pt
    \cellcolor[HTML]{FFBC98}\rawvalue %
  \else\ifdim \rawvalue pt < 64pt
    \cellcolor[HTML]{FFC3A2}\rawvalue %
  \else\ifdim \rawvalue pt < 66pt
    \cellcolor[HTML]{FFCAAC}\rawvalue %
  \else\ifdim \rawvalue pt < 68pt
    \cellcolor[HTML]{FFD1B6}\rawvalue %
  \else\ifdim \rawvalue pt < 70pt
    \cellcolor[HTML]{FFD8C0}\rawvalue %
  \else\ifdim \rawvalue pt < 72pt
    \cellcolor[HTML]{FFDFCA}\rawvalue %
  \else\ifdim \rawvalue pt < 74pt
    \cellcolor[HTML]{FFE6D4}\rawvalue %
  \else\ifdim \rawvalue pt < 76pt
    \cellcolor[HTML]{FFEDDE}\rawvalue %
  \else\ifdim \rawvalue pt < 78pt
    \cellcolor[HTML]{FFF4E8}\rawvalue %
  \else\ifdim \rawvalue pt < 80pt
    \cellcolor[HTML]{FFFFF2}\rawvalue %
  \else\ifdim \rawvalue pt < 82pt
    \cellcolor[HTML]{F2FFF2}\rawvalue %
  \else\ifdim \rawvalue pt < 84pt
    \cellcolor[HTML]{E5FFE5}\rawvalue %
  \else\ifdim \rawvalue pt < 86pt
    \cellcolor[HTML]{D8FFD8}\rawvalue %
  \else\ifdim \rawvalue pt < 88pt
    \cellcolor[HTML]{CBFFCB}\rawvalue %
  \else\ifdim \rawvalue pt < 90pt
    \cellcolor[HTML]{BEFFBE}\rawvalue %
  \else\ifdim \rawvalue pt < 92pt
    \cellcolor[HTML]{B1FFB1}\rawvalue %
  \else\ifdim \rawvalue pt < 94pt
    \cellcolor[HTML]{A4FFA4}\rawvalue %
  \else\ifdim \rawvalue pt < 96pt
    \cellcolor[HTML]{97FF97}\rawvalue %
  \else\ifdim \rawvalue pt < 98pt
    \cellcolor[HTML]{8AFF8A}\rawvalue %
  \else\ifdim \rawvalue pt < 100pt
    \cellcolor[HTML]{7DFF7D}\rawvalue %
  \else\ifdim \rawvalue pt < 101pt
    \cellcolor[HTML]{70FF70}\rawvalue %
  \else
    \rawvalue %
  \fi\fi\fi\fi\fi\fi\fi\fi\fi\fi\fi\fi\fi\fi\fi\fi\fi\fi\fi\fi\fi\fi\fi\fi\fi\fi
  \endgroup
}

\def\argmax{\qopname\relax n{argmax}}

\title{WiC is Not WSD: A Study on LLMs and Lexical Ambiguity Resolution}

\author{
        Yi Zhou$^{1*}$ \quad
        Kiamehr Rezaee$^{1*}$ \quad
        Danushka Bollegala$^{2,3}$ \\
        \textbf{Mohammad Taher Pilehvar}$^{1}$ \quad
        \textbf{Jose Camacho-Collados}$^{1}$\\
        $^1$Cardiff University\quad
        $^2$University of Liverpool \quad
        $^3$Amazon\\
        \tt{\{zhouy131, RezaeeK, CamachoColladosJ, PilehvarMT\}@cardiff.ac.uk} \\
        \quad
        {\tt danushka@liverpool.ac.uk}
}

\date{}

\begin{document}
\maketitle
\def\thefootnote{*}\footnotetext{Equal contribution.}

\begin{abstract}
Word-in-Context (WiC) remains challenging for language models, despite recent progress on lexical-semantic tasks. We hypothesise that this difficulty arises not only from comparing two contextual uses of a word, but also from the absence of an explicit sense inventory that specifies the relevant level of semantic granularity. We evaluate open LLMs on WiC and traditional Word Sense Disambiguation (WSD) under similar settings. We find that providing candidate senses, similar to what is done in traditional WSD, improves WiC performance in all settings. In general, explicit sense information helps models make more consistent and targeted judgements. Human evaluation further shows that many apparent WiC errors reflect label ambiguity or mismatches between model and annotator sense boundaries rather than simple failures of lexical understanding. In particular, results show that LLMs overthink the sense distinction often leading to errors based on overly fine-grained distinctions.

\end{abstract}


\section{Introduction}





Lexical ambiguity has long been considered a fundamental problem in NLP. 
A single word form may correspond to multiple meanings, and understanding language therefore requires identifying the intended meaning of a word in context. 
For decades, the Word Sense Disambiguation (WSD) task has provided a standard formulation of this problem. 
Given a target word in context and a predefined inventory of candidate senses, a model must select the sense that best matches the intended meaning~\cite{navigli2009word}.
Although the rise of contextualised embedding methods and Large Language Models (LLMs) has changed how lexical meaning is represented in language models, lexical ambiguity remains an important diagnostic for evaluating whether model semantic distinctions align with human-interpretable senses.

\begin{figure}[t!]
    \centering
    \includegraphics[width=\linewidth]{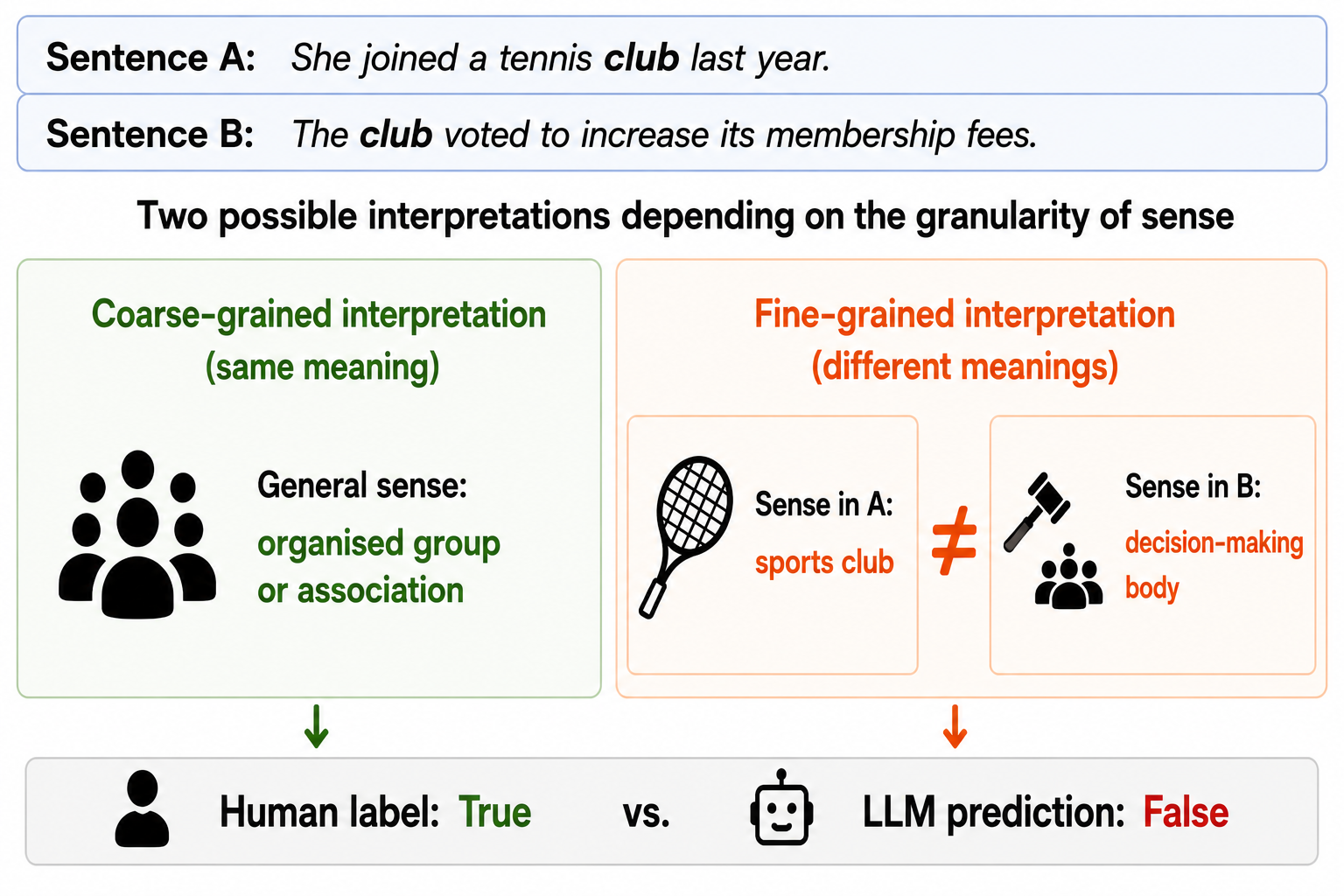}

    \caption{Example of an LLM error on a WiC instance caused by different assumptions about sense granularity. At a coarse-grained level, both uses of \textit{club} refer to an organised group or association, leading to the human label \texttt{True}. The LLM instead adopts a finer-grained distinction between the two sentences, leading to the prediction \texttt{False}.}

    \label{fig:first_image}
    \vspace{-0.5em}

\end{figure}


The Word in Context (WiC) task was introduced as an evaluation of context-sensitive meaning representations~\cite{pilehvar-camacho-collados-2019-wic}, and has been part of language understanding benchmarks such as SuperGLUE \cite{wang2019superglue}. 
Unlike WSD, WiC does not require a model to select a sense from an explicit sense inventory. 
Instead, given two sentences containing the same target word, a model must decide whether the word is used in the same meaning in both contexts. 
Despite its simple binary format, WiC has been difficult for LLMs, with GPT-3 and Chinchilla reported to perform close to random chance in prompted WiC settings~\cite{brown2020language,wei2022emergent}. 
This raises important questions about the nature of the task itself: What makes WiC so challenging for LLMs? Is it merely the presentation of word usage across two different sentences, or does the challenge stem from more fundamental aspects of how semantic understanding is evaluated?

We argue that the difficulty of WiC lies not only in the comparison of two independent contexts, but also in the absence of an explicitly specified level of semantic granularity. 
In WSD, the candidate sense inventory specifies the set of distinctions that the model is expected to make. 
The model is instructed to select from a predefined set of alternatives, where the sense inventory determines the granularity at which the decision is evaluated.
In contrast, in WiC, the model must judge whether two contextualised uses of a word take the same or different meanings without giving an explicit inventory or a criterion for how fine-grained the distinction should be.
As a result, WiC requires both contextual interpretation and implicit alignment with the sense distinctions assumed by the dataset annotators. Figure~\ref{fig:first_image} shows an example of this issue with the word \textit{club}.
At a coarse level of granularity, both instances of \emph{club} can be considered as referring to an organised group or an association, thus the expected WiC label will be \texttt{True}.
However, an LLM may focus on finer contextual differences and distinguishes a sporting or social club from an administrative decision-making body, and therefore predict \texttt{False}.

In this paper, we show that when sense granularity information is explicitly provided to models in a WiC-like format, performance improves substantially, particularly for smaller models. This improvement is also found if we explicitly require models to solve WiC through a preliminary WSD step. These results suggest that the difficulty models experience with the original WiC formulation stems primarily from the ambiguity in determining appropriate sense distinctions rather than an inability to compare contextual word meanings.


Finally, we present evidence to disentangle the sources of errors made by various models on the WiC formatted task. Specifically, we distinguish between: (1) errors resulting from a model’s inability to correctly identify the meaning of the target word in each sentence, (2) errors arising from the failure to perform the minimal required reasoning step in the form of a comparison between the inferred meanings, and (3) errors arising from a mismatch between the models' preferences for sense distinction and those of human annotators who labelled the dataset.

\section{Related Work}
WSD has been studied through a wide range of benchmarks and shared tasks. 
\citet{raganato-etal-2017-word} introduce a unified framework that brings together several standard WSD datasets and enables systematic comparison between models. 
An influential line of work uses glosses or sense definitions as part of the input. 
\citet{huang-etal-2019-glossbert} pair each target context with candidate glosses and use BERT~\cite{devlin-etal-2019-bert} to estimate their compatibility, while \citet{blevins-zettlemoyer-2020-moving} show that gloss-informed bi-encoders are particularly helpful for rare senses. 

More recently, WSD has been revisited in the context of LLMs.
\citet{basile2025exploring} evaluate LLMs on WSD-style tasks in which models either generate the correct definition or select the correct meaning from a predefined set.
They find that LLMs perform well in zero-shot settings, although they do not surpass current state-of-the-art WSD systems.
Other work has explored fine-tuning and prompting LLMs for WSD~\cite{paev-etal-2025-word,sumanathilaka2025glossgpt}.

WiC has also been extended and reused in more recent work as a diagnostic task for contextual lexical meaning. 
\citet{martelli-etal-2021-semeval} introduce a multilingual version of WiC, designed to evaluate whether systems can distinguish word senses both within and between languages without relying on a fixed sense inventory.
More recently, WiC has been used to probe lexical semantic knowledge in LLMs. 
\citet{hayashi-2024-reassessing} use WiC to reassess the semantic knowledge encoded in LLMs, focusing on whether models can judge semantic equivalence across contexts.
\citet{hayashi-2025-evaluating} further evaluate LLMs on lexical semantic equivalence using WiC-style prompting with both proprietary and open-source models.

\citet{hauer-kondrak-2022-wic} analyse the formal relationship between WiC, WSD, and Target Sense Verification~(TSV;~\citealp{breit2021wic}), a task in which a model is given a target word in context together with a candidate sense and must decide whether the candidate sense matches the intended meaning of the word. 
They argue that the three tasks can be reduced to one another under the assumption that distinctions in meaning correspond to distinctions between senses. 
This result clarifies why resources and systems developed for WSD can often be adapted to WiC or TSV.
However, formal reducibility does not necessarily mean that tasks are equivalent to evaluation settings for language models, as we show in this paper.

Finally, recent work by \citet{mujko-etal-2026-insights} shows that models trained in WSD can transfer effectively to WiC, and that models trained in WiC can also improve WSD performance. 
In addition, they find that joint training can benefit WiC, especially in low-resource settings. 
These results suggest that WiC and WSD draw on related abilities for sense interpretation but that their relationship depends on how the tasks are formulated and evaluated. 

\section{Background}
\label{sec:background}
\subsection{Word Sense Disambiguation (WSD)}
WSD is the task of determining the most appropriate meaning of a word based on its context.
In its standard formulation, WSD uses an external sense inventory, such as WordNet~\cite{miller1995wordnet,fellbaum1998wordnet} listing the possible senses of a target word.

More formally, let \( w \) be a target word that occurs in a context \( c \), and let \( S(w) = \{s_1, s_2, \dots, s_n\} \) denote the set of candidate senses for \( w \), as defined by an external lexical resource such as WordNet.
The goal of WSD is to identify the correct sense \( s^* \in S(w) \) that best matches the intended meaning of \( w \) in context \( c \). Formally, this can be framed as a candidate selection problem:
\begin{align}
    s^* = \argmax_{s_i \in S(w)} P(s_i \mid w, c),
\end{align}
where \( P(s_i \mid w, c) \) denotes the probability that the sense \( s_i \) is the correct interpretation of \( w \) given the surrounding context \( c \).
A well performing WSD system must model the semantic distinctions among all candidates in \( S(w) \) and assign the highest probability to the sense that best fits the context.
Alternatively, the level of granularity for candidate senses \( S(w) \) can be adjusted to suit the specific use case by providing the model with suitable coarse-grained options.

\subsection{Word-in-Context (WiC)}

Unlike WSD, WiC does not provide the model with a candidate sense inventory.
Instead, the task is to determine whether the meaning of a given target word is similar in two different contexts.

Formally, given two contexts \( c_1 \) and \( c_2 \), both containing an instance of a target word \( w \), the goal is to determine whether the sense of \( w \) in \( c_1 \) is the same as in \( c_2 \). The model must output a label \( y \in \{\texttt{True}, \texttt{False}\} \), where:
\begin{align}
    y = 
\begin{cases}
\texttt{True} & \text{if } \text{sense}(w, c_1) = \text{sense}(w, c_2) \\
\texttt{False} & \text{otherwise}
\end{cases}
\end{align}

A fundamental distinction is that WiC lacks an explicit sense inventory \( S(w) \). The set of possible meanings is implicit in the distribution of examples within the dataset and in the judgments of human annotators. In contrast, WSD relies on a fixed inventory \( S(w) \).
This structural difference means that, in WiC, the model must internally determine the appropriate sense granularity without explicit guidance, while in WSD, the granularity is externally defined and enforced. As a result, WiC serves as a test of how well the model’s latent sense representations align with human judgments, rather than its ability to select from a known inventory.




\section{Experimental Setting}

In light of the fundamental differences between the WiC and WSD tasks discussed in Section~\ref{sec:background}, we reassess the ability of LLMs to resolve lexical ambiguity when explicit sense options are provided.

\subsection{Models}

Regarding the choice of models, we select a range of openly available LLMs, including small and large variants of the \textbf{LLaMA3}~\cite{grattafiori2024llama}, \textbf{Mistral}~\cite{jiang2023mistral}, and \textbf{DeepSeek}~\cite{bi2024deepseek} models. This selection allows us to analyse how model size, pretraining objectives, and architectural differences influence performance. The specific models used throughout this paper are listed in Table \ref{tab:models}.

\begin{table}[h]
    \resizebox{\linewidth}{!}{

\begin{tabular}{llll}
\toprule
Model & Variant & Repository & Size \\
\midrule
Llama3 & L & \href{https://huggingface.co/meta-llama/Meta-Llama-3-70B-Instruct}{Meta-Llama-3-70B-Instruct} & 70B \\
Llama3 & S & \href{https://huggingface.co/meta-llama/Meta-Llama-3-8B-Instruct}{Meta-Llama-3-8B-Instruct} & 8B \\
DeepSeek & L & \href{https://huggingface.co/deepseek-ai/DeepSeek-R1-Distill-Llama-70B}{DeepSeek-R1-Distill-Llama-70B} & 70B \\
DeepSeek & S & \href{https://huggingface.co/deepseek-ai/DeepSeek-R1-Distill-Llama-8B}{DeepSeek-R1-Distill-Llama-8B} & 8B \\
Mistral & L & \href{https://huggingface.co/mistralai/Mixtral-8x7B-Instruct-v0.1}{Mixtral-8x7B-Instruct-v0.1} & 8x7B \\
Mistral & S & \href{https://huggingface.co/mistralai/Mistral-7B-Instruct-v0.2}{Mistral-7B-Instruct-v0.2} & 7B \\
\bottomrule
\end{tabular}

}
    \caption{Models used in our experiments. L and S denote the large and small variants within each model family, respectively. Repository names are linked to the corresponding Hugging Face model pages.}
    \label{tab:models}
    \vspace{-0.5em}
\end{table}

\subsection{Tasks and Datasets}

We conduct experiments to evaluate model performance across three different settings: (1) WiC with coarse-grained sense distinctions, (2) WiC with fine-grained senses, and (3) WSD with coarse-grained senses. 
Comparing the results in these tasks offers valuable insight into the sources of errors and limitations in the current LLMs’ ability of handling lexical ambiguity.

For the coarse-grained WSD task, we use the \textbf{CWSD-20} dataset~\cite{loureiro2021analysis}, which is a WSD dataset that contains sentences centered around 20 ambiguous target words. In each sentence, a target word is used in context with one of its coarse-grained senses.
Moreover, to create a coarse-grained version for WiC style evaluation, we automatically convert CWSD-20 into WiC format. 
We process the instances in CWSD-20 by pairing sentences containing the same target word and assigning a binary label indicating semantic similarity.
We first extract labels and corresponding sentences from CWSD-20, where each distinct meaning of a target word is represented by a class label. 
We then compare the class label of each pair of consecutive sentences. 
If the target word in both sentences has the same class label (i.e., with the same meaning), the pair is labeled as \texttt{True}, otherwise, it is labeled as \texttt{False}.

For the fine-grained WiC setting, we introduce a new dataset based on how the original WiC benchmark was constructed~\cite{pilehvar-camacho-collados-2019-wic}. We decided not to use the original benchmark given potential contamination issues as the dataset has been widely available by LLMs. Instead, we introduced modifications intended to limit the influence of memorisation by language models. 
To this end, we extract examples from WordNet 3.1, focusing on noun and verb entries. 
Specifically, we extract target words and their associated definitions by parsing structured WordNet instances.
We associate each target word with its corresponding gloss, creating a mapping between words and their sense definitions to provide contextual grounding.
Next, we build WiC-style sentence pairs by identifying multiple usage examples for a given sense and constructing all valid example pairs in which the same target word appears in both sentences.

The summary statistics of the coarse-grained WiC, fine-grained WiC and coarse-grained WSD datasets we used are shown in Table~\ref{tab:overall-statistics}. 
Examples and detailed statistics for the coarse-grained WiC, fine-grained WiC and coarse-grained WSD datasets are provided in Appendices~\ref{app:coarse-grained-wic-dataset}, \ref{app:fine-grained-wic-dataset} and~\ref{app:cwsd20-sample-statistics}, respectively.

\begin{table}[t]
\centering
\resizebox{\linewidth}{!}{
\begin{tabular}{lrrr}
\toprule
Statistic & Coarse WiC & Fine WiC & Coarse WSD \\
\midrule
Instances & 2,798 & 1,185 & 33,566 \\
Target words & 20 & 924 & 20 \\
Distinct senses & 53 & 1,655 & 53 \\
Unique sentences & 5,594 & 2,266 & 33,566 \\
\texttt{True} labels & 1,399 & 594 & -- \\
\texttt{False} labels & 1,399 & 591 & -- \\
\bottomrule
\end{tabular}}
\caption{Summary statistics of the datasets used in the experiments. Distinct senses refers to coarse-grained sense labels for Coarse WiC and Coarse WSD, and to WordNet sense definitions for Fine WiC.}
\label{tab:overall-statistics}
\vspace{-0.5em}
\end{table}


\subsection{Prompting Strategies}

A standard approach for evaluating LLMs is prompt-based evaluation, which we adopt for all our experiments.

For the WSD task, the prompt presents the model with a sentence, a target word, and a list of candidate senses. The model is asked to choose the candidate that best matches the meaning of the target word in the given context.

For the WiC task, the prompt includes two sentences containing the same target word, along with the target word itself. The model is then asked to answer \texttt{True} or \texttt{False} depending on whether it believes the word has the same meaning in both sentences. Unlike WSD, the standard WiC formulation does not provide explicit candidate senses. However, as discussed earlier, the absence of defined sense distinctions may leave the task underspecified, since the intended granularity of sense differentiation is not always clear to the model. To address this, we also conduct experiments using an extended version of WiC where candidate senses are explicitly provided. Throughout this paper, we refer to the standard format (without candidates) as \textbf{Options $-$}, and the extended format (with candidates) as \textbf{Options $+$}.

Additionally, we also investigate the effect of chain-of-thought (CoT) prompting, where models are asked to provide an explanation before giving the final answer. For both WiC and WSD, we run experiments with and without CoT. Experiments in which models are instructed to provide an explanation before giving their final answer are labelled \textbf{CoT $+$}, whereas those without explanation are denoted \textbf{CoT $-$}. Examples of the prompts used in each setting are provided in Appendix~\ref{app:prompt}.

\begin{table*}[!ht]
\centering
\resizebox{\linewidth}{!}{
\begin{tabular}{lllrrrrrrr}
\toprule
\multirow{2}{*}{Task} & \multicolumn{2}{c}{Setting} & \multicolumn{6}{c}{Model} & \multirow{2}{*}{Mean} \\
\cmidrule(lr){2-3} \cmidrule(lr){4-9}
& Options & CoT & Llama3-L & Llama3-S & DeepSeek-L & DeepSeek-S & Mistral-L & Mistral-S &  \\
\midrule
\multirow{4}{*}{Coarse WiC}
& + & + & \textbf{95.3} & \textbf{83.5} & \textbf{96.1} & \textbf{88.7} & \textbf{84.2} & 61.5 & \textbf{84.9} \\
& + & -- & 93.9 & 75.3 & 94.1 & 86.6 & 72.8 & 63.1 & 81.0 \\
& -- & + & 85.9 & 67.0 & 89.3 & 80.8 & 74.8 & \textbf{73.0} & 78.5 \\
& -- & -- & 85.5 & 65.9 & 86.3 & 83.2 & 66.2 & 58.0 & 74.9 \\
\midrule
\multirow{4}{*}{Fine WiC}
& + & + & 79.3 & \textbf{66.6} & 77.4 & 74.8 & \textbf{70.1} & 59.9 & 71.4 \\
& + & -- & \textbf{80.4} & 64.1 & \textbf{77.6} & \textbf{76.4} & 69.5 & 63.5 & \textbf{71.9} \\
& -- & + & 71.1 & 56.2 & 75.2 & 68.3 & 66.4 & \textbf{63.7} & 66.8 \\
& -- & -- & 75.1 & 59.1 & 73.5 & 69.9 & 68.6 & 59.7 & 67.7 \\
\midrule
\multirow{2}{*}{Coarse WSD}
& + & + & \textbf{97.0} & \textbf{90.1} & \textbf{95.2} & \textbf{93.7} & \textbf{91.4} & \textbf{89.2} & \textbf{92.7} \\
& + & -- & 96.6 & 83.3 & 94.0 & 92.8 & 82.0 & 68.8 & 86.3 \\
\bottomrule
\end{tabular}}
\caption{Average accuracy across the three evaluation settings. Results are averaged over 20 target words for the coarse-grained WiC and coarse-grained WSD tasks, and reported on the newly constructed fine-grained WiC dataset. Coarse and fine-grained WiC are evaluated with and without candidate sense options, while coarse-grained WSD always provides a candidate sense inventory. Options+ indicates that candidate senses are provided, and CoT+ indicates that chain-of-thought prompting is used. Bold marks the best score for each model within each task.}
\label{tab:main-results}
\end{table*}

\section{Results}
Table \ref{tab:main-results} reports the average results for coarse-grained WiC, fine-grained WiC, and coarse-grained WSD. 
For the coarse-grained WiC and WSD tasks, results are averaged over the 20 target words.
Detailed per-word results are provided in Tables \ref{tab:detailed_wic_opt+}, \ref{tab:detailed_wic_opt-} and \ref{tab:detailed_wsd} in Appendix~\ref{app:full_results}. 




\paragraph{Effect of Candidates Sense Options.}

Although the standard task formulation in WiC does not include a list of candidate senses, our results demonstrate that providing a list of sense options clearly enhances the model performance by giving the model direct access to the appropriate sense granularity for each instance, rather than requiring it to infer this information independently. 
In the coarse-grained WiC setting, the highest mean performance is obtained when both options and CoT prompting are provided, reaching an average of 84.9\% over all models. 
When CoT is kept fixed, the addition of candidate options improves the mean score from 78.5\% to 84.9\% in the CoT+ setting and from 74.9\% to 81.0\% in the CoT- setting. 
This indicates that candidate senses provide useful information beyond the surface context themselves. 

A similar trend appears in the fine-grained WiC setting. 
Although overall performance is lower than in the coarse-grained setting, the configurations with candidate options still achieve the strongest mean scores. 
The best mean result in the fine-grained task is 71.9\%, obtained with Options+ and CoT-. 
By contrast, the corresponding Options- setting obtains 67.7\%.
These results suggest that explicit sense information is useful even when the relevant distinctions are fine-grained and more difficult to apply. 

The improvement from Options+ supports our hypothesis that WiC difficulty is partly caused by the absence of an explicit sense inventory. 
Providing sense options makes the task becomes closer to WSD or target sense verification, because the model is no longer required to infer the relevant level of sense granularity entirely on its own. 
However, this does not make WiC identical to WSD, since the model still has to compare the two contextual uses.
We note that the improvement from providing candidate senses should not be attribute to sense granularity alone, since the candidates also provide additional semantic information about the target word. 
We therefore consider this result together with the analysis in Section~\ref{sec:performance_analysis}, where we compare WiC and WSD under the same sense inventory and examine the types of errors made by LLMs. 


\paragraph{Effect of Chain-of-Thought Prompting.}





CoT prompting improves average performance in the coarse-grained WiC setting, but not in the fine-grained setting. 
In coarse-grained WiC, CoT improves the mean score in both the Options+ and Options- settings. 
With options, the mean score rises from 81.0\% to 84.9\%, while without options, it rises from 74.9\% to 78.5\%.
This suggests that explicit reasoning can help models compare contextual meaning, especially when the sense distinctions are relatively coarse. 

However, the fine-grained WiC results show a different pattern. 
In this settings, CoT does not consistently improve performance. 
Specifically, the mean score drops from 71.9\% without CoT to 71.4\% with CoT.
The same trend appears when no options are provided, where the mean score drops from 67.7\% without CoT to 66.8\% with CoT. 

This may be because in the fine-grained Options- setting, the model has no explicit sense inventory to anchor the comparison. 
When asked to explain its reasoning, the model often focuses on the contextual details that distinguish the two sentences. 
In fine-grained tasks, these details may reflect differences in usage or emphasis, but do not always indicate a different annotated sense.  
CoT can therefore amplify small contextual differences and encourage the model to draw overly fine-grained boundaries between related uses. 
This is consistent with our human evaluation results in Section~\ref{sec:performance_analysis}, where granularity-related errors are frequently observed.



\paragraph{Model Family and Model Size.}
Across the coarse-grained WiC task, DeepSeek and Llama3 models generally perform better than Mistral models. 
In the best-performing configuration (i.e., Options+ and CoT+), DeepSeek-L achieves 96.1\%, Llama3-L achieves 95.3\%, and Mistral-L achieves 84.2\%.
A similar pattern appears in the coarse-grained WSD results with CoT+, where Llama3-L achieves 97.0\%, DeepSeek-L achieves 95.2\%, and Mistral-L reaches 91.4\%.

The size of the model also plays an substantial role.
Large models generally outperform their smaller variants in both WiC and WSD settings. 
The gap is especially large for Mistral, where the small model performs poorly in several configurations. 
However, the small DeepSeek model is a notable exception to the general size trend. 
In both coarse-grained WiC and WSD, DeepSeek-S achieves better performance compared to Mistral-L.
This suggests that the size of the model is important, but it is not the only factor. 
Model design also appear to affect lexical-semantic performance.

\paragraph{Impact of Sense Granularity.}

When comparing the fine-grained and coarse-grained WiC tasks, it is evident that models perform better when the sense distinctions are more easily distinguishable, reflecting the selection criteria used in constructing coarse-grained WiC sentences. In contrast, performance drops in the fine-grained setting, suggesting that overly subtle sense distinctions pose a greater challenge for all models. 
This aligns with human intuitions, since excessively fine-grained senses are also known to be difficult for humans to reliably differentiate~\cite{hovy-etal-2006-ontonotes, palmer-dang-fellbaum-2007}. 

\begin{figure*}[!htbp]

        \includegraphics[width=\linewidth]{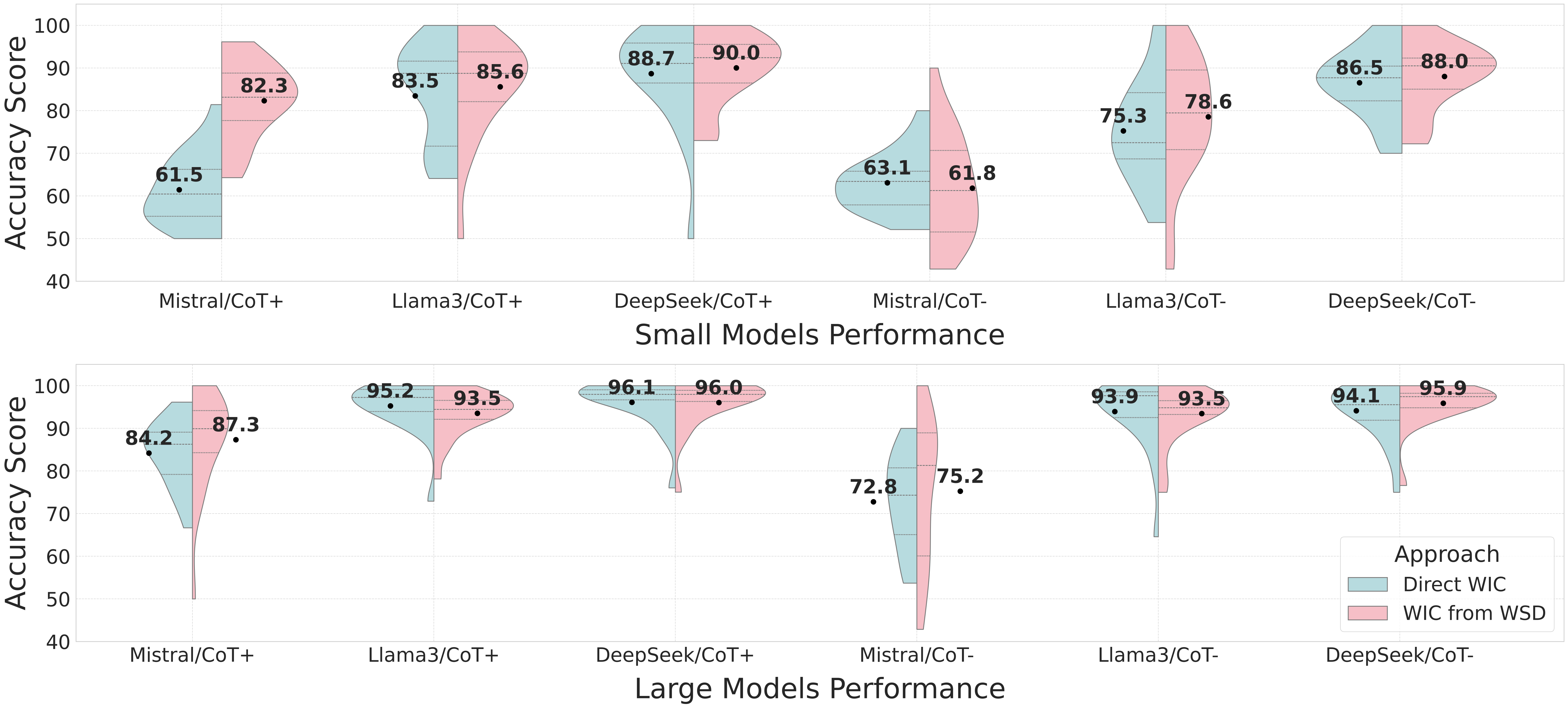}
        
    \caption{Comparison between direct WiC prediction and WiC prediction through WSD on the coarse-grained setting. Violin plots show the distribution of accuracy scores across the 20 target words for each model and prompting condition. The top panel reports small-model performance and the bottom panel reports large-model performance. Numbers indicate mean accuracy.}
    \label{fig:wic_vs_wsd2wic}

\end{figure*}

\section{Analysis of Performance}
\label{sec:performance_analysis}
The experimental results show that explicit sense information improves WiC performance, but they do not explain where the remaining errors come from. 
An incorrect WiC prediction may reflect several different issues.
For example, the model may fail to identify the sense of the target word in one or both contexts, fail to compare two inferred meanings correctly, or adopt a sense boundary that differs from the one assumed by the dataset.
In this section, we analyse these possibilities.

\paragraph{Lexical semantic knowledge of models.}
One strategy to solve a WiC instance is to first determine the sense of the target word in each sentence, and then compare the two inferred senses.
The WSD results provide a direct assessment of whether models can recognise coarse-grained word senses when the relevant candidate inventory is provided. 
As shown by the coarse-grained WSD results in Table~\ref{tab:main-results}, models achieve a mean score of 92.7\% with CoT prompting, with individual scores ranging from 89.2\% to 97.0.
Even the smaller models perform competitively in this setting, with Llama3-S reaching 90.1\%, DeepSeek-S reaching 93.7\% and Mistral-S reaching 89.16\%.
This suggests that the models have enough lexical-semantic knowledge to select the intended coarse-grained sense in a single context. 

However, this does not directly translate into the same level of WiC performance. 
In the corresponding coarse-grained WiC setting with candidate options and CoT, Table~\ref{tab:main-results} shows a mean score of 84.9\%, compared with 92.7\% for WSD.
Importantly, the same sense inventory is available in both settings.
The gap therefore suggests that WiC adds difficulty beyond single-context sense selection. 
Unlike WSD, WiC requires the model to judge whether two contextual uses should be treated as the same meaning under the dataset's intended sense granularity. 
This motivates the WiC through WSD analysis, where we test whether making the sense selection explicit improves WiC decisions. 


\paragraph{Solving WiC through WSD.}
\label{wsd2wic}
We therefore introduce a WiC through WSD setting in which the model resolves each target word occurrence separately before the two predicted senses are compared. 
In this setting, the model first selects the sense of the target word in each sentence from an explicit candidate inventory. 
The final WiC label is then derived from the two WSD predictions by an identity comparison. 
If the two predicted senses are identical, the label is \texttt{True}, otherwise, it is \texttt{False}.
This keeps the final WiC decision unchanged, but makes the intermediate sense selection step explicit. We use our coarsed-grained WiC dataset for this task, since we have the sense distinctions for each dataset based on the original CWSD-20 WSD dataset. 

Figure~\ref{fig:wic_vs_wsd2wic} shows that WiC through WSD is especially helpful for smaller models. 
For Mistral-S with CoT, the mean score increases from 61.5\% in direct to 82.3\% with WiC through WSD.
Smaller gains are also observed for Llama3-S and DeepSeek-S, which improve from 83.5\% to 85.6\% and from 88.7\% to 90.0\% under CoT, respectively. 
For larger models, the effect is weaker and less consistent. 
Mistral-L improves from 84.2\% to 87.3\% with CoT, but Llama3-L decreases from 95.2\% to 93.5\%, and DeepSeek-L remains almost unchanged, moving from 96.1\% to 96.0\%.

These results suggest that decomposing WiC into explicit WSD followed by sense comparison mainly helps models that cannot reliably perform this decomposition implicitly. 
The clear improvement for smaller models supports our claim that part of the WiC challenging nature is due to the absence of an explicit sense inventory and granularity criterion. 
When LLMs are required to select the senses explicitly before making the WiC judgment, performance improves particularly for smaller models, suggesting that making the relevant sense distinctions explicit can help models align their comparison with intended sense boundaries.

\paragraph{Bias Toward \texttt{False} Predictions}
Table~\ref{tab:false-bias} shows that in the coarse-grained WiC setting, the models predict \texttt{False} more often than \texttt{True} in every configuration. 
The \texttt{False} prediction rate ranges from 60.8\% with Options+ and CoT+ to 69.1\% with Options- and CoT-.
This pattern is driven mainly by low recall on \texttt{True} instances. 
While recall on gold \texttt{False} instances stays between 93.5\% and 95.4\%, recall on gold \texttt{True} instances ranges from 73.8\% to 55.8\%.

The pattern is especially relevant for gold \texttt{True} instances, where two contextual uses are expected to be grouped under the same sense. 
A \texttt{False} prediction in such cases does not necessarily mean that the model misunderstands either occurrence. 
Instead, the model may be applying a finer-grained sense distinction than the dataset assumes. 
The human evaluation results support this interpretation.

\begin{table}[t!]
\setlength{\tabcolsep}{10pt}
\resizebox{\linewidth}{!}{
\begin{tabular}{llrrr}
\toprule
Options & CoT & Pred$_{\texttt{F}}$ & Recall$_{\texttt{F}}$ & Recall$_{\texttt{T}}$ \\
\midrule
$+$ & $+$ & 60.8 & 95.4 & 73.8 \\
$-$ & $+$ & 63.6 & 93.5 & 66.4 \\
$+$ & $-$ & 65.7 & 94.3 & 62.8 \\
$-$ & $-$ & 69.1 & 94.0 & 55.8 \\
\bottomrule
\end{tabular}}
\caption{Class-specific behaviour on the coarse-grained WiC task, averaged across models. Pred$_{\texttt{F}}$ indicates the percentage of predictions assigned to \texttt{False}. Recall$_{\texttt{F}}$ and Recall$_{\texttt{T}}$ report recall for each gold class.}
\label{tab:false-bias}
\end{table}

\begin{table}[!t]
\resizebox{\linewidth}{!}{
\begin{tabular}{lrr}
\toprule
Category & Coarse WiC & Fine WiC \\
\midrule
Unique annotated instances & 120 & 140 \\
Total annotations & 186 & 188 \\
\midrule
Human label \texttt{True} & 86 & 84 \\
Human label \texttt{False} & 34 & 47 \\

\midrule
Model judged correct & 39 & 76 \\
Sense misidentification & 54 & 19 \\
Wrong final decision & 21 & 33 \\
Other & 6 & 3 \\
Missing or undecided & 0 & 9 \\
\midrule
Overly fine-grained distinction & 32 & 31 \\
Overly coarse-grained distinction & 0 & 14 \\
\bottomrule
\end{tabular}}
\caption{Human annotation results for sampled model errors. Counts are computed over unique instances using majority labels when multiple annotations are available. ``Overly fine-grained distinction'' means that the model applies a finer boundary than the human annotation, separating two uses that the human annotation treats as the same sense, while ``Overly coarse-grained distinction'' means that the model applies a coarser sense boundary than then human annotation, groups two uses that the human annotation treats as different senses.}
\label{tab:human-error-analysis}
\vspace{-0.6em}
\end{table}

\paragraph{Human Evaluation of Errors}
\label{pa:human_evalua}
To better understand the sources of error in model predictions on the WiC task, we conducted a qualitative analysis by sampling model errors across both fine-grained as well as coarse-grained task variants.

For the fine-grained WiC, a total of 140 erroneous model predictions were randomly selected. Of these, 40 instances were annotated independently by two annotators, while the remaining samples were annotated by one of them. Similarly, for the coarse-grained WiC task, 120 samples were selected, with 40 double annotated and the rest annotated by a single annotator.

Annotators were first presented with the original WiC instance (two sentences containing the target word, along with the word itself) and were asked to label whether the meanings of the target word in the two contexts were the same or different (effectively re-annotating the instance to detect any potential label noise). Next, they were shown the model’s response to the same instance and asked to choose one of the following judgments: (1) The model is correct and not making any mistakes; (2) The model assigns incorrect meanings to one or both sentences; (3) The model correctly identifies the meanings of the target word in each sentence but arrives at an incorrect final answer (i.e., it chooses the wrong label); or (4) other (for instances that are unclear, incomplete or otherwise do not fit the previous categories). 
For responses judged to involve an incorrect final answer, annotators were also asked to indicate whether the model used an overly fine-grained or overly coarse-grained sense distinction. 
An incorrect final answer may reflect a mismatch in granularity between the model’s sense interpretation and the gold label or a failure in performing the simple reasoning step required to compare the two senses.
All samples were drawn from the configuration in which CoT prompting was enabled, allowing annotators to observe the model’s reasoning process. Sampling was balanced across model families and sizes to ensure broad coverage.

Table~\ref{tab:human-error-analysis} summarizes the results of the human evaluation. 
In the coarse-grained setting, sense misidentification is the largest category, accounting for 54 out of 120 instances. 
However, 39 instances are judged to be correct despite being sampled as model errors under the original dataset label. 
This suggests that a substantial portion of apparent errors may reflect label ambiguity or disagreement about sense boundaries rather than model failure. 
In the fine-grained setting, this pattern is even stronger. 
Annotators judge that model to be correct in 76 out of 140 instances, while only 19 instances are classified as sense misidentification errors. 
This indicates that fine-grained WiC contains more cases where the model prediction is linguistically plausible even when it disagree with the dataset label. 

Granularity-related annotations further support this interpretation. 
In the coarse-grained setting, 32 instances are marked as overly fine-grained distinction, while no instance is marked as overly coarse-grained. 
In the fine-grained setting, both types appear, with 31 overly fine-grained and 14 overly coarse-grained cases. 
These results suggest that WiC errors are not always failures of lexical-semantic knowledge.
They often reflect a mismatch between the preferred sense boundary of models and the granularity assumed by the dataset. 
In particular, models tend to overthink potential fine-grained sense distinctions rather than assessing senses as being the same.
In many of these cases, an LLM identifies a contextually plausible meaning but draws a sense boundary that differs from human annotators. 
The difficulty therefore does not always arise from failing to understand the contextual meaning itself, but also from determining which semantic distinctions should matter for the WiC judgment.










\section{Conclusion}

This paper investigated why WiC remains difficult for language models. 
We argued that challenge is not only that WiC requires comparing two contexts, but also that it leaves the relevant level of sense granularity unspecified. 
Unlike WSD, where the candidate sense inventory make the intended distinctions explicit, WiC requires models to decide how fine-grained the comparison should be. 
This causes LLMs to often come up with overly fine-grained distinctions compared to the gold standard.

Our findings show that providing candidate sense options improves perforamnce in both coarse-grained and fine-grained WiC settings. 
More importantly, WiC remains more difficult than WSD even when the same sense inventory is provided. 
The human analysis further shows that many errors arise from differences between the sense boundaries adopted by the models and those used in the annotations. 
In addition, explicitly selecting senses before making the WiC judgement improves performance for several models. 
Overall, our results suggest that determining and applying the appropriate level of semantic distinction is an important contributing factor to WiC difficulty.  
In addition, we show that larger models are generally better at handling sense distinctions, while smaller models benefit more from explicit sense guidance. Further research would be required to fully understand the implications of this finding, and to develop techniques to induce this capability to smaller models.

\section*{Acknowledgements}
Danushka Bollegala holds concurrent appointments as a Professor at University of Liverpool and as an Amazon Scholar. 
This paper describes work performed at the University of Liverpool and is not associated with Amazon.

\section*{Limitations}

Our experiments focus on English lexical ambiguity. 
This allows for a controlled comparison between WiC, WSD, and WiC-through-WSD, but future work should examine whether the same patterns hold in multilingual settings. 
This is especially relevant because sense boundaries may differ across languages, and the same lexical distinction may not always be preserved in translation.

We evaluate a selected set of open LLMs from several model families and sizes. 
This selection is sufficient to show clear differences between model families and between smaller and larger models, but it does not cover the full range of available language models. 

The coarse-grained experiments are based on 20 target words, which enables detailed per-word analysis and controlled comparison across settings. 
However, lexical ambiguity is a broad phenomenon, and other types of ambiguity may behave differently. 
Future work could extend the analysis to larger inventories and to cases involving metaphorical, idiomatic, domain-specific, or discourse-dependent meaning shifts.

Providing candidate sense options changes the original WiC formulation, making it closer to WSD or TSV. 
This is intentional in our experimental design, since our goal is to isolate the effect of explicit sense granularity on WiC-style decisions. 
The results should therefore be interpreted as evidence about the role of explicit sense information, rather than as a replacement for the standard WiC benchmark.

Finally, our human evaluation is based on sampled model errors. 
This sampling allows us to inspect error types in detail and identify cases of label ambiguity and granularity mismatch. 
A larger annotation study would provide a more complete estimate of the frequency of each error type, and could further clarify how human and model sense boundaries diverge.

\section{Ethical Considerations}

This work uses existing lexical-semantic resources and evaluates open language models on word-sense understanding tasks. The experiments do not involve personal, sensitive, or user-generated private data. The human annotation component focuses only on model outputs and sentence-level lexical ambiguity examples, and annotators were asked to judge semantic relations and error types rather than provide personal information.

\bibliography{acl2021}


\appendix

\section*{Appendix}

\section{Dataset Details}
\label{app:dataset-details}
This appendix provides additional details about the datasets used in our experiments. 
Appendix~\ref{app:coarse-grained-wic-dataset} describes the coarse-grained WiC dataset derived from CWSD-20, Appendix~\ref{app:fine-grained-wic-dataset} describes the newly constructed fine-grained WiC dataset, and Appendix~\ref{app:cwsd20-sample-statistics} reports examples and statistics for the original coarse-grained WSD dataset (CWSD-20).

\begin{table}[t]
\centering
\small
\begin{tabular}{lrrrr}
\toprule
Target word & No. senses & Pairs & \texttt{True} & \texttt{False} \\
\midrule
\textit{apple} & 2 & 424 & 212 & 212 \\
\textit{arm} & 2 & 70 & 35 & 35 \\
\textit{bank} & 2 & 46 & 23 & 23 \\
\textit{bass} & 3 & 508 & 254 & 254 \\
\textit{bow} & 3 & 106 & 53 & 53 \\
\textit{chair} & 2 & 40 & 20 & 20 \\
\textit{club} & 3 & 96 & 48 & 48 \\
\textit{crane} & 2 & 78 & 39 & 39 \\
\textit{deck} & 2 & 14 & 7 & 7 \\
\textit{digit} & 2 & 10 & 5 & 5 \\
\textit{hood} & 3 & 38 & 19 & 19 \\
\textit{java} & 2 & 824 & 412 & 412 \\
\textit{mole} & 5 & 70 & 35 & 35 \\
\textit{pitcher} & 2 & 26 & 13 & 13 \\
\textit{pound} & 2 & 20 & 10 & 10 \\
\textit{seal} & 4 & 108 & 54 & 54 \\
\textit{spring} & 3 & 182 & 91 & 91 \\
\textit{square} & 4 & 82 & 41 & 41 \\
\textit{trunk} & 3 & 38 & 19 & 19 \\
\textit{yard} & 2 & 18 & 9 & 9 \\
\midrule
Total & 53 & 2,798 & 1,399 & 1,399 \\
\bottomrule
\end{tabular}
\caption{Per-word statistics of the coarse-grained WiC dataset constructed from CWSD-20.}
\label{tab:coarse-wic-statistics}
\end{table}

\begin{table*}[h!]
\centering
\small
\begin{tabular}{p{0.09\linewidth}p{0.36\linewidth}p{0.36\linewidth}p{0.08\linewidth}}
\toprule
Target & Sentence 1 & Sentence 2 & Label \\
\midrule
\textit{bank} 
& She deposited the cheque at the \textit{bank}. 
& The customer opened an account at the \textit{bank}. 
& \texttt{True} \\

\textit{bank} 
& The boat stopped near the river \textit{bank}. 
& She deposited the cheque at the \textit{bank}. 
& \texttt{False} \\

\textit{club} 
& She joined a tennis \textit{club} last year. 
& The \textit{club} voted to increase its membership fees. 
& \texttt{True} \\

\textit{club} 
& He carried a heavy wooden \textit{club}. 
& She joined a tennis \textit{club} last year. 
& \texttt{False} \\

\textit{spring} 
& The water flows from a natural \textit{spring}. 
& The village was built around a fresh-water \textit{spring}. 
& \texttt{True} \\

\textit{spring} 
& The water flows from a natural \textit{spring}. 
& The \textit{spring} was unusually warm this year. 
& \texttt{False} \\
\bottomrule
\end{tabular}
\caption{Illustrative examples from the coarse-grained WiC dataset. Each instance contains one target word, two contexts, and a binary label indicating whether the target word has the same sense in both sentences.}
\label{tab:coarse-wic-examples}
\end{table*}

\subsection{Coarse-Grained WiC Dataset}
\label{app:coarse-grained-wic-dataset}

The coarse-grained WiC dataset is constructed from CWSD-20 by pairing sentences that contain the same target word and comparing their coarse-grained sense labels. 
If the two sentences have the same sense label, the WiC label is \texttt{True}; otherwise, it is \texttt{False}. 
Table~\ref{tab:coarse-wic-statistics} reports the per-word statistics of the resulting dataset. 
The dataset contains 2,798 sentence pairs across 20 target words, with a balanced label distribution of 1,399 \texttt{True} and 1,399 \texttt{False} instances. 
Table~\ref{tab:coarse-wic-examples} shows representative examples.

\subsection{Fine-Grained WiC Dataset}
\label{app:fine-grained-wic-dataset}

Table~\ref{tab:wic-dataset-statistics} shows the statistics of the fine-grained WiC-style dataset used in our experiments. The dataset contains 1,185 sentence pairs covering 924 unique target words. 
Each instance consists of a target word, two contexts, two corresponding sense definitions, and a binary label indicating whether the two uses express the same meaning. The label distribution is nearly balanced, with 594 \texttt{True} instances and 591 \texttt{False} instances. 

Table~\ref{tab:wic-dataset-examples} shows representative examples from the dataset. 
\texttt{True} instances contain two uses corresponding to the same WordNet sense, while \texttt{False} instances contain two uses corresponding to different senses. 

\begin{table}[h]
\centering
\begin{tabular}{lr}
\toprule
Statistic & Value \\
\midrule
Instances & 1,185 \\
Unique target words & 924 \\
Unique definitions & 1,655 \\
Unique sentence pairs & 1,185 \\
Unique sentences & 2,266 \\
\texttt{True} labels & 594 \\
\texttt{False} labels & 591 \\
\texttt{True} / \texttt{False} ratio & 50.13 / 49.87 \\
\bottomrule
\end{tabular}
\caption{Statistics of the fine-grained WiC dataset.}
\label{tab:wic-dataset-statistics}
\end{table}

\begin{table*}[h]
\centering
\resizebox{\textwidth}{!}{
\begin{tabular}{llllll}
\toprule
Target & Sentence 1 & Sentence 2 & Definition 1 & Definition 2 & Label \\
\midrule
\textit{tar} & tar the roads & tar the roof & coat with tar & coat with tar & \texttt{T} \\
\textit{leaf} & leaf a manuscript & leaf through a book & turn over pages & turn over pages & \texttt{T} \\
\textit{room} & room for improvement & room to pass & opportunity for & space for movement & \texttt{F} \\
\textit{hand} & give me a hand with the chores & he wanted to try his hand at singing & physical assistance & ability & \texttt{F} \\
\textit{voice} & he wanted to hear trained voices sing it & he lost his voice & a singer & the ability to speak & \texttt{F} \\
\bottomrule
\end{tabular}
}
\caption{Example instances from the fine-grained WiC dataset with sense information. Each instance contains a target word, two contexts, two sense definitions, and a binary label.}
\label{tab:wic-dataset-examples}
\end{table*}

\subsection{Coarse WSD Dataset (CWSD-20)}
\label{app:cwsd20-sample-statistics}

Table~\ref{tab:coarsewsd-statistics} reports statistics for CoarseWSD-20. 
The dataset contains 20 ambiguous nouns and 53 coarse-grained senses in total. 
Each target word is associated with a separate classification dataset, with sentences split into training and test sets using a 60/40 split. 
The statistics in the table are computed from the sense frequencies reported.

Table~\ref{tab:coarsewsd-samples} shows representative examples of CoarseWSD-style instances. 
Each instance contains a target word, a context sentence, and the corresponding coarse-grained sense. 
The examples illustrate the type of human-interpretable sense distinctions used in the dataset.

\begin{table}[h]
\centering
\resizebox{\linewidth}{!}{
\begin{tabular}{lrrrr}
\toprule
Target word & No. senses & Train & Test & Total \\
\midrule
\textit{apple} & 2 & 2,358 & 1,032 & 3,390 \\
\textit{arm} & 2 & 423 & 164 & 587 \\
\textit{bank} & 2 & 1,107 & 455 & 1,562 \\
\textit{bass} & 3 & 3,173 & 1,391 & 4,564 \\
\textit{bow} & 3 & 523 & 215 & 738 \\
\textit{chair} & 2 & 271 & 130 & 401 \\
\textit{club} & 3 & 388 & 202 & 590 \\
\textit{crane} & 2 & 372 & 157 & 529 \\
\textit{deck} & 2 & 170 & 99 & 269 \\
\textit{digit} & 2 & 68 & 42 & 110 \\
\textit{hood} & 3 & 171 & 82 & 253 \\
\textit{java} & 2 & 4,504 & 1,929 & 6,433 \\
\textit{mole} & 5 & 480 & 206 & 686 \\
\textit{pitcher} & 2 & 6,421 & 2,819 & 9,240 \\
\textit{pound} & 2 & 186 & 97 & 283 \\
\textit{seal} & 4 & 875 & 363 & 1,238 \\
\textit{spring} & 3 & 1,064 & 457 & 1,521 \\
\textit{square} & 4 & 508 & 207 & 715 \\
\textit{trunk} & 3 & 164 & 77 & 241 \\
\textit{yard} & 2 & 144 & 72 & 216 \\
\midrule
Total & 53 & 23,370 & 10,196 & 33,566 \\
\bottomrule
\end{tabular}}
\caption{Statistics of CoarseWSD-20. The dataset contains 20 target words and 53 coarse-grained senses. Train and test counts are obtained by summing the per-sense frequencies reported.}
\label{tab:coarsewsd-statistics}
\end{table}

\begin{table}[h]
\resizebox{\linewidth}{!}{
\begin{tabular}{p{0.13\linewidth}p{0.55\linewidth}p{0.24\linewidth}}
\toprule
Target & Context & Sense \\
\midrule
\multirow{2}{*}{\textit{bank}} 
& She deposited the cheque at the \textit{bank}. 
& financial institution \\
& The children played near the river \textit{bank}. 
& river side \\
\midrule

\multirow{2}{*}{\textit{club}} 
& She joined a tennis \textit{club} last year. 
& organisation or association \\
& He carried a heavy wooden \textit{club}. 
& weapon \\
\midrule

\multirow{2}{*}{\textit{spring}} 
& The water flows from a natural \textit{spring}. 
& water source \\
& The \textit{spring} was unusually warm this year. 
& season \\
\bottomrule
\end{tabular}}
\caption{Illustrative CoarseWSD-style examples. The target word is shown in italics.}
\label{tab:coarsewsd-samples}
\end{table}

\section{Full Results}
Tables~\ref{tab:detailed_wic_opt+} and~\ref{tab:detailed_wic_opt-} report the coarse-grained WiC results with and without candidate sense options, respectively. For each setting, results are shown separately for CoT+ and CoT- prompting across all evaluated models. Table~\ref{tab:detailed_wsd} reports the corresponding per-word results for the coarse-grained WSD task.
\label{app:full_results}

\begin{table*}[!ht]
    \resizebox{\textwidth}{!}{\begin{tabular}{llrrrrrr}
\toprule
 & \multirow{2}{*}{Word} & \multicolumn{6}{c}{Model} \\
\cmidrule(lr){3-8}
 & & Llama3-L & Llama3-S & DeepSeek-L & DeepSeek-S & Mistral-L & Mistral-S \\
\midrule
\multirow{21}{*}{\rotatebox{90}{CoT+}}
& apple   & 97.88 & 68.40 & 98.58 & 88.68 & 79.72 & 50.47 \\
& arm     & 98.57 & 88.57 & 98.57 & 95.71 & 87.14 & 81.43 \\
& bank    & 97.83 & 71.74 & 97.83 & 93.48 & 93.48 & 60.87 \\
& bass    & 89.37 & 75.00 & 87.99 & 82.09 & 77.56 & 63.39 \\
& bow     & 99.06 & 90.57 & 98.11 & 97.17 & 87.74 & 55.66 \\
& chair   & 95.00 & 87.50 & 95.00 & 95.00 & 82.50 & 72.50 \\
& club    & 72.92 & 64.58 & 76.04 & 73.96 & 66.67 & 54.17 \\
& crane   & 94.87 & 64.10 & 98.72 & 89.74 & 89.74 & 56.41 \\
& deck    & 92.86 & 71.43 & 100.00 & 50.00 & 71.43 & 71.43 \\
& digit   & 100.00 & 100.00 & 100.00 & 100.00 & 90.00 & 60.00 \\
& hood    & 92.11 & 94.74 & 97.37 & 92.11 & 86.84 & 65.79 \\
& java    & 99.27 & 96.60 & 98.67 & 96.24 & 88.59 & 57.04 \\
& mole    & 94.29 & 90.00 & 92.86 & 90.00 & 82.86 & 62.86 \\
& pitcher & 100.00 & 96.15 & 100.00 & 100.00 & 96.15 & 53.85 \\
& pound   & 95.00 & 65.00 & 90.00 & 75.00 & 75.00 & 50.00 \\
& seal    & 97.22 & 88.89 & 97.22 & 88.89 & 83.33 & 53.70 \\
& spring  & 97.25 & 89.01 & 97.80 & 93.41 & 85.71 & 67.58 \\
& square  & 91.46 & 89.02 & 97.56 & 85.37 & 75.61 & 70.73 \\
& trunk   & 100.00 & 94.74 & 100.00 & 86.84 & 94.74 & 65.79 \\
& yard    & 100.00 & 83.33 & 100.00 & 100.00 & 88.89 & 55.56 \\
\cmidrule(lr){2-8}
& \textbf{Average} & 95.25 & 83.47 & 96.12 & 88.68 & 84.18 & 61.46 \\
\midrule
\multirow{21}{*}{\rotatebox{90}{CoT-}}
& apple   & 95.05 & 53.77 & 97.41 & 81.37 & 60.38 & 52.12 \\
& arm     & 98.57 & 85.71 & 95.71 & 90.00 & 82.86 & 64.29 \\
& bank    & 97.83 & 58.70 & 100.00 & 82.61 & 67.39 & 58.70 \\
& bass    & 87.01 & 72.44 & 82.87 & 80.12 & 77.17 & 58.07 \\
& bow     & 98.11 & 81.13 & 98.11 & 88.68 & 79.25 & 60.38 \\
& chair   & 90.00 & 72.50 & 92.50 & 95.00 & 75.00 & 65.00 \\
& club    & 64.58 & 69.79 & 75.00 & 75.00 & 54.17 & 62.50 \\
& crane   & 97.44 & 62.82 & 100.00 & 89.74 & 65.38 & 58.97 \\
& deck    & 92.86 & 78.57 & 85.71 & 71.43 & 85.71 & 57.14 \\
& digit   & 100.00 & 100.00 & 100.00 & 100.00 & 90.00 & 80.00 \\
& hood    & 100.00 & 84.21 & 94.74 & 86.84 & 78.95 & 65.79 \\
& java    & 98.91 & 72.21 & 95.15 & 91.75 & 64.08 & 55.10 \\
& mole    & 98.57 & 85.71 & 90.00 & 88.57 & 80.00 & 68.57 \\
& pitcher & 100.00 & 100.00 & 100.00 & 100.00 & 88.46 & 65.38 \\
& pound   & 80.00 & 60.00 & 90.00 & 70.00 & 70.00 & 65.00 \\
& seal    & 98.15 & 71.30 & 95.37 & 85.19 & 53.70 & 57.41 \\
& spring  & 98.35 & 65.38 & 96.15 & 95.05 & 73.63 & 72.53 \\
& square  & 91.46 & 74.39 & 93.90 & 86.59 & 69.51 & 73.17 \\
& trunk   & 97.37 & 84.21 & 100.00 & 89.47 & 84.21 & 65.79 \\
& yard    & 94.44 & 72.22 & 100.00 & 83.33 & 55.56 & 55.56 \\
\cmidrule(lr){2-8}
& \textbf{Average} & 93.94 & 75.25 & 94.13 & 86.56 & 72.77 & 63.07 \\
\bottomrule
\end{tabular}}
    \caption{Per-word accuracy on the coarse-grained WiC task with candidate sense options provided. Results are reported for each model under CoT+ and CoT- settings.}
    \label{tab:detailed_wic_opt+}
\end{table*}

\begin{table*}[!ht]
    \resizebox{\textwidth}{!}{\begin{tabular}{llrrrrrr}
\toprule
 & \multirow{2}{*}{Word} & \multicolumn{6}{c}{Model} \\
\cmidrule(lr){3-8}
 & & Llama3-L & Llama3-S & DeepSeek-L & DeepSeek-S & Mistral-L & Mistral-S \\
\midrule
\multirow{21}{*}{\rotatebox{90}{CoT+}}
& apple   & 86.32 & 58.73 & 88.44 & 82.78 & 76.89 & 66.98 \\
& arm     & 92.86 & 70.00 & 88.57 & 88.57 & 81.43 & 81.43 \\
& bank    & 91.30 & 60.87 & 95.65 & 82.61 & 67.39 & 71.74 \\
& bass    & 83.86 & 63.19 & 83.07 & 75.39 & 78.35 & 67.91 \\
& bow     & 93.40 & 71.70 & 94.34 & 91.51 & 84.91 & 76.42 \\
& chair   & 77.50 & 62.50 & 97.50 & 80.00 & 77.50 & 67.50 \\
& club    & 58.33 & 59.38 & 67.71 & 59.38 & 56.25 & 56.25 \\
& crane   & 89.74 & 66.67 & 91.03 & 78.21 & 80.77 & 75.64 \\
& deck    & 78.57 & 57.14 & 64.29 & 71.43 & 71.43 & 85.71 \\
& digit   & 70.00 & 40.00 & 90.00 & 70.00 & 60.00 & 70.00 \\
& hood    & 68.42 & 73.68 & 81.58 & 78.95 & 73.68 & 76.32 \\
& java    & 95.02 & 74.15 & 95.51 & 90.05 & 86.04 & 74.39 \\
& mole    & 82.86 & 74.29 & 84.29 & 80.00 & 80.00 & 74.29 \\
& pitcher & 100.00 & 84.62 & 100.00 & 96.15 & 84.62 & 92.31 \\
& pound   & 100.00 & 60.00 & 95.00 & 75.00 & 75.00 & 60.00 \\
& seal    & 84.26 & 73.15 & 93.52 & 80.56 & 68.52 & 65.74 \\
& spring  & 93.96 & 69.23 & 92.31 & 88.46 & 74.73 & 81.32 \\
& square  & 85.37 & 73.17 & 91.46 & 73.17 & 67.07 & 73.17 \\
& trunk   & 97.37 & 81.58 & 97.37 & 84.21 & 73.68 & 71.05 \\
& yard    & 88.89 & 66.67 & 94.44 & 88.89 & 77.78 & 72.22 \\
\cmidrule(lr){2-8}
& \textbf{Average} & 85.90 & 67.04 & 89.30 & 80.77 & 74.80 & 73.02 \\
\midrule
\multirow{21}{*}{\rotatebox{90}{CoT-}}
& apple   & 75.94 & 55.42 & 89.39 & 80.19 & 64.62 & 56.37 \\
& arm     & 90.00 & 78.57 & 87.14 & 91.43 & 74.29 & 64.29 \\
& bank    & 91.30 & 54.35 & 95.65 & 76.09 & 58.70 & 50.00 \\
& bass    & 78.74 & 67.72 & 83.07 & 74.80 & 72.83 & 57.48 \\
& bow     & 89.62 & 73.58 & 93.40 & 92.45 & 70.75 & 52.83 \\
& chair   & 82.50 & 52.50 & 90.00 & 92.50 & 65.00 & 57.50 \\
& club    & 67.71 & 58.33 & 65.62 & 63.54 & 56.25 & 55.21 \\
& crane   & 89.74 & 70.51 & 89.74 & 79.49 & 73.08 & 62.82 \\
& deck    & 78.57 & 71.43 & 78.57 & 78.57 & 71.43 & 57.14 \\
& digit   & 80.00 & 60.00 & 70.00 & 70.00 & 60.00 & 60.00 \\
& hood    & 84.21 & 65.79 & 73.68 & 81.58 & 73.68 & 60.53 \\
& java    & 93.69 & 71.48 & 93.69 & 88.96 & 66.38 & 59.34 \\
& mole    & 90.00 & 70.00 & 84.29 & 74.29 & 71.43 & 62.86 \\
& pitcher & 100.00 & 84.62 & 100.00 & 96.15 & 73.08 & 61.54 \\
& pound   & 65.00 & 50.00 & 80.00 & 85.00 & 55.00 & 55.00 \\
& seal    & 89.81 & 69.44 & 87.96 & 83.33 & 59.26 & 57.41 \\
& spring  & 90.11 & 64.84 & 90.11 & 86.81 & 67.03 & 63.19 \\
& square  & 90.24 & 70.73 & 85.37 & 82.93 & 64.63 & 54.88 \\
& trunk   & 94.74 & 73.68 & 94.74 & 84.21 & 65.79 & 60.53 \\
& yard    & 88.89 & 55.56 & 94.44 & 100.00 & 61.11 & 50.00 \\
\cmidrule(lr){2-8}
& \textbf{Average} & 85.54 & 65.93 & 86.34 & 83.16 & 66.22 & 57.95 \\
\bottomrule
\end{tabular}}
    \caption{Per-word accuracy on the coarse-grained WiC task with no candidate sense options provided. Results are reported for each model under CoT+ and CoT- settings.}
    \label{tab:detailed_wic_opt-}
\end{table*}

\begin{table*}[!ht]
    \resizebox{\textwidth}{!}{\begin{tabular}{llrrrrrr}
\toprule
 & \multirow{2}{*}{Word} & \multicolumn{6}{c}{Model} \\
\cmidrule(lr){3-8}
 & & Llama3-L & Llama3-S & DeepSeek-L & DeepSeek-S & Mistral-L & Mistral-S \\
\midrule
\multirow{21}{*}{\rotatebox{90}{CoT+}}
& apple   & 99.41 & 80.78 & 96.70 & 96.79 & 88.68 & 91.27 \\
& arm     & 100.00 & 98.57 & 99.15 & 97.13 & 89.29 & 92.14 \\
& bank    & 100.00 & 95.56 & 95.32 & 96.27 & 97.78 & 96.67 \\
& bass    & 90.94 & 78.25 & 89.90 & 87.32 & 76.18 & 50.79 \\
& bow     & 99.53 & 97.17 & 97.30 & 94.12 & 97.17 & 96.23 \\
& chair   & 97.50 & 97.50 & 94.50 & 96.50 & 93.75 & 96.25 \\
& club    & 81.77 & 79.69 & 82.54 & 81.43 & 78.65 & 77.08 \\
& crane   & 97.44 & 97.44 & 94.92 & 95.25 & 94.87 & 86.54 \\
& deck    & 96.43 & 46.43 & 96.43 & 88.37 & 75.00 & 85.71 \\
& digit   & 100.00 & 95.00 & 100.00 & 100.00 & 100.00 & 85.00 \\
& hood    & 96.05 & 94.74 & 94.70 & 93.81 & 89.47 & 97.37 \\
& java    & 99.82 & 99.45 & 94.65 & 94.82 & 94.60 & 91.38 \\
& mole    & 97.14 & 91.43 & 93.29 & 93.16 & 95.71 & 95.71 \\
& pitcher & 100.00 & 100.00 & 99.08 & 98.45 & 100.00 & 100.00 \\
& pound   & 95.00 & 92.50 & 94.50 & 92.72 & 90.00 & 87.50 \\
& seal    & 97.69 & 85.19 & 96.06 & 93.07 & 78.70 & 74.07 \\
& spring  & 97.80 & 92.31 & 96.40 & 92.21 & 94.78 & 95.33 \\
& square  & 98.17 & 88.41 & 94.95 & 91.43 & 95.12 & 90.85 \\
& trunk   & 97.37 & 96.05 & 96.25 & 94.24 & 98.68 & 96.05 \\
& yard    & 97.22 & 94.44 & 96.40 & 96.36 & 100.00 & 97.22 \\
\cmidrule(lr){2-8}
& \textbf{Average} & 96.96 & 90.05 & 95.15 & 93.67 & 91.42 & 89.16 \\
\midrule
\multirow{21}{*}{\rotatebox{90}{CoT-}}
& apple   & 98.82 & 45.05 & 96.58 & 95.99 & 51.53 & 45.75 \\
& arm     & 100.00 & 97.14 & 99.29 & 96.43 & 56.43 & 56.43 \\
& bank    & 98.89 & 96.67 & 94.44 & 96.67 & 97.78 & 82.22 \\
& bass    & 90.65 & 64.96 & 87.60 & 85.73 & 60.73 & 33.46 \\
& bow     & 100.00 & 96.70 & 96.70 & 93.40 & 96.70 & 87.26 \\
& chair   & 97.50 & 87.50 & 92.50 & 96.25 & 70.00 & 66.25 \\
& club    & 77.60 & 77.60 & 82.81 & 79.17 & 69.27 & 60.42 \\
& crane   & 98.72 & 98.72 & 92.95 & 95.51 & 89.10 & 61.54 \\
& deck    & 92.86 & 53.57 & 96.43 & 78.57 & 64.29 & 57.14 \\
& digit   & 100.00 & 85.00 & 100.00 & 100.00 & 90.00 & 75.00 \\
& hood    & 96.05 & 90.79 & 93.42 & 92.11 & 90.79 & 88.16 \\
& java    & 99.70 & 85.01 & 95.08 & 95.87 & 75.00 & 59.22 \\
& mole    & 97.14 & 88.57 & 89.29 & 92.86 & 97.86 & 75.71 \\
& pitcher & 100.00 & 100.00 & 98.08 & 98.08 & 100.00 & 94.23 \\
& pound   & 95.00 & 70.00 & 87.50 & 90.00 & 80.00 & 72.50 \\
& seal    & 97.69 & 90.28 & 93.06 & 91.67 & 84.72 & 46.76 \\
& spring  & 97.53 & 75.55 & 96.43 & 94.51 & 94.51 & 84.34 \\
& square  & 97.56 & 74.39 & 96.95 & 90.85 & 90.85 & 70.12 \\
& trunk   & 98.68 & 94.74 & 96.05 & 94.74 & 86.84 & 84.21 \\
& yard    & 97.22 & 94.44 & 94.44 & 97.22 & 94.44 & 75.00 \\
\cmidrule(lr){2-8}
& \textbf{Average} & 96.58 & 83.33 & 93.98 & 92.78 & 82.04 & 68.79 \\
\bottomrule
\end{tabular}}
    \caption{Per-word accuracy on the coarse-grained WSD task. Results are reported for each model under CoT+ and CoT- settings.}
    \label{tab:detailed_wsd}
\end{table*}

\section{Prompt Templates}
\label{app:prompt}

We use three prompt families in the experiments, corresponding to coarse-grained WiC, fine-grained WiC, and coarse-grained WSD. 
For the WiC tasks, the model is given two sentences containing the same target word and is asked to decide whether the target word has the same meaning in both contexts. 
For the WSD task, the model is given one sentence and asked to select the meaning of the target word from a predefined list of candidate senses.

For both coarse-grained and fine-grained WiC, we evaluate two task formats. 
In the Options+ format, the prompt includes a list of candidate meanings or definitions for the target word. 
This gives the model explicit access to the relevant sense inventory and the intended level of sense granularity. 
In the Options- format, no candidate meanings are provided, and the model must make the same-or-different judgement directly from the two contexts. 
For the coarse-grained WSD task, candidate meanings are always provided, since selecting from an explicit inventory is part of the WSD formulation.

We also evaluate two prompting strategies. 
In the CoT+ setting, the model is asked to first state the meaning of the target word in each context, or to identify the best candidate meaning, before producing the final answer. 
In the CoT- setting, the model is instructed to provide only the final answer without explanation. 
All prompts require the final response to follow a fixed format, either \texttt{Answer: [[True]]} or \texttt{Answer: [[False]]} for WiC, and \texttt{Answer: [[label]]} for WSD. Tables~\ref{tab:prompt-coarse-wic}, \ref{tab:prompt-fine-wic}, and \ref{tab:prompt-coarse-wsd} show the prompt templates used for coarse-grained WiC, fine-grained WiC, and coarse-grained WSD, respectively.

\begin{table*}[t]
\centering
\small
\begin{tabular}{p{0.14\linewidth}p{0.80\linewidth}}
\toprule
Setting & Prompt template \\
\midrule

Options+ / CoT+ 
&
\begin{minipage}[t]{0.80\linewidth}
\ttfamily
You will be given two sentences and your task is to determine whether the meaning of the target word \{word\} is similar in the two sentences or not. 
In each sentence, the meaning of the word \{word\} refers to one of the following possible meanings: meaning 1 - \{meaning\_1\}; meaning 2 - \{meaning\_2\}; ... meaning n - \{meaning\_n\}. 
You need to first provide which of the possible meanings from the provided candidates suits the target word \{word\} in each sentence best. 
Then provide a Boolean label, that is `True' if the meaning of \{word\} is the same in the two sentences, or `False' if it is not. 
Your final Boolean response should strictly follow this format: `Answer: [[label]]', where the label in the double brackets is either True or False, e.g., `Answer: [[True]]' or `Answer: [[False]]'. 

Target Word: `\{word\}' 

Sentence 1: \{sentence\_1\} 

Sentence 2: \{sentence\_2\}
\end{minipage}
\\
\midrule

Options+ / CoT- 
&
\begin{minipage}[t]{0.80\linewidth}
\ttfamily
You will be given two sentences and your task is to determine whether the meaning of the target word \{word\} is similar in the two sentences or not. 
In each sentence, the meaning of the word \{word\} refers to one of the following possible meanings: meaning 1 - \{meaning\_1\}; meaning 2 - \{meaning\_2\}; ... meaning n - \{meaning\_n\}. 
You need to provide a Boolean label, and you should not provide any sort of explanation of your thinking process. 
The label is `True' if the meaning of \{word\} is the same in the two sentences, or `False' if it is not. 
Your final Boolean response should strictly follow this format: `Answer: [[label]]', where the label in the double brackets is either True or False, e.g., `Answer: [[True]]' or `Answer: [[False]]'. 

Target Word: `\{word\}' 

Sentence 1: \{sentence\_1\} 

Sentence 2: \{sentence\_2\}
\end{minipage}
\\
\midrule

Options- / CoT+ 
&
\begin{minipage}[t]{0.80\linewidth}
\ttfamily
You will be given two sentences and your task is to determine whether the target word \{word\} refers to the same meaning in the two sentences or not. 
You need to first provide what the meaning of the target word \{word\} is in each sentence. 
Then provide a Boolean label, that is `True' if the meaning of \{word\} is the same in the two sentences, or `False' if it is not. 
Your final Boolean response should strictly follow this format: `Answer: [[label]]', where the label in the double brackets is either True or False, e.g., `Answer: [[True]]' or `Answer: [[False]]'. 

Target Word: `\{word\}' 

Sentence 1: \{sentence\_1\} 

Sentence 2: \{sentence\_2\}
\end{minipage}
\\
\midrule

Options- / CoT- 
&
\begin{minipage}[t]{0.80\linewidth}
\ttfamily
You will be given two sentences and your task is to determine whether the target word \{word\} refers to the same meaning in the two sentences or not. 
You need to provide a Boolean label, and you should not provide any sort of explanation of your thinking process. 
The label is `True' if the meaning of \{word\} is the same in the two sentences, or `False' if it is not. 
Your final Boolean response should strictly follow this format: `Answer: [[label]]', where the label in the double brackets is either True or False, e.g., `Answer: [[True]]' or `Answer: [[False]]'. 

Target Word: `\{word\}' 

Sentence 1: \{sentence\_1\} 

Sentence 2: \{sentence\_2\}
\end{minipage}
\\

\bottomrule
\end{tabular}
\caption{Prompt templates used for the coarse-grained WiC experiments. Options+ indicates that candidate sense meanings are provided, while Options- indicates that no candidate meanings are provided. CoT+ prompts the model to state the meaning of the target word in each sentence before giving the final label, while CoT- requests only the final label.}
\label{tab:prompt-coarse-wic}
\end{table*}

\begin{table*}[t]
\centering
\small
\begin{tabular}{p{0.14\linewidth}p{0.80\linewidth}}
\toprule
Setting & Prompt template \\
\midrule

Options+ / CoT+ 
&
\begin{minipage}[t]{0.80\linewidth}
\ttfamily
You will be given two sentences and your task is to determine whether the meaning of the target word \{word\} is similar in the two sentences or not. 
In each sentence, the meaning of the word \{word\} refers to one of the following possible definitions: definition 1 - \{definition\_1\}; definition 2 - \{definition\_2\}; ... definition n - \{definition\_n\}. 
You need to first provide which of the possible definitions from the provided candidates suits the target word \{word\} in each sentence best. 
Then provide a Boolean label, that is `True' if the meaning of \{word\} is the same in the two sentences, or `False' if it is not. 
Your final Boolean response should strictly follow this format: `Answer: [[label]]', where the label in the double brackets is either True or False, e.g., `Answer: [[True]]' or `Answer: [[False]]'. 

Target Word: `\{word\}' 

Sentence 1: \{sentence\_1\} 

Sentence 2: \{sentence\_2\}
\end{minipage}
\\
\midrule

Options+ / CoT- 
&
\begin{minipage}[t]{0.80\linewidth}
\ttfamily
You will be given two sentences and your task is to determine whether the meaning of the target word \{word\} is similar in the two sentences or not. 
In each sentence, the meaning of the word \{word\} refers to one of the following possible definitions: definition 1 - \{definition\_1\}; definition 2 - \{definition\_2\}; ... definition n - \{definition\_n\}. 
You need to provide a Boolean label, and you should not provide any sort of explanation of your thinking process. 
The label is `True' if the meaning of \{word\} is the same in the two sentences, or `False' if it is not. 
Your final Boolean response should strictly follow this format: `Answer: [[label]]', where the label in the double brackets is either True or False, e.g., `Answer: [[True]]' or `Answer: [[False]]'. 

Target Word: `\{word\}' 

Sentence 1: \{sentence\_1\} 

Sentence 2: \{sentence\_2\}
\end{minipage}
\\
\midrule

Options- / CoT+ 
&
\begin{minipage}[t]{0.80\linewidth}
\ttfamily
You will be given two sentences and your task is to determine whether the target word \{word\} refers to the same meaning in the two sentences or not. 
You need to first provide what the meaning of the target word \{word\} is in each sentence. 
Then provide a Boolean label, that is `True' if the meaning of \{word\} is the same in the two sentences, or `False' if it is not. 
Your final Boolean response should strictly follow this format: `Answer: [[label]]', where the label in the double brackets is either True or False, e.g., `Answer: [[True]]' or `Answer: [[False]]'. 

Target Word: `\{word\}' 

Sentence 1: \{sentence\_1\} 

Sentence 2: \{sentence\_2\}
\end{minipage}
\\
\midrule

Options- / CoT- 
&
\begin{minipage}[t]{0.80\linewidth}
\ttfamily
You will be given two sentences and your task is to determine whether the target word \{word\} refers to the same meaning in the two sentences or not. 
You need to provide a Boolean label, and you should not provide any sort of explanation of your thinking process. 
The label is `True' if the meaning of \{word\} is the same in the two sentences, or `False' if it is not. 
Your final Boolean response should strictly follow this format: `Answer: [[label]]', where the label in the double brackets is either True or False, e.g., `Answer: [[True]]' or `Answer: [[False]]'. 

Target Word: `\{word\}' 

Sentence 1: \{sentence\_1\} 

Sentence 2: \{sentence\_2\}
\end{minipage}
\\

\bottomrule
\end{tabular}
\caption{Prompt templates used for the fine-grained WiC experiments. Options+ indicates that candidate WordNet-style definitions are provided, while Options- indicates that no candidate definitions are provided. CoT+ prompts the model to identify the meaning of the target word in each sentence before giving the final label, while CoT- requests only the final label.}
\label{tab:prompt-fine-wic}
\end{table*}

\begin{table*}[t]
\centering
\small
\begin{tabular}{p{0.14\linewidth}p{0.80\linewidth}}
\toprule
Setting & Prompt template \\
\midrule

CoT+ 
&
\begin{minipage}[t]{0.80\linewidth}
\ttfamily
You will be given a sentence and your task is to determine the meaning of the target word \{word\} from the following options: meaning 1 - \{meaning\_1\}; meaning 2 - \{meaning\_2\}; ... meaning n - \{meaning\_n\}. 
You need to first provide which of the possible meanings from the provided candidates suits the target word \{word\} in the given sentence best. 
Then provide the numerical label. 
You have to provide the explanation first, this step is necessary. 
After you provide your explanation, your final response should be the numerical label strictly following this format: `Answer: [[label]]', where the label in the double brackets is the numerical label from the provided list, e.g., `Answer: [[1]]'. 

Target Word: `\{word\}' 

Sentence: \{sentence\}
\end{minipage}
\\
\midrule

CoT- 
&
\begin{minipage}[t]{0.80\linewidth}
\ttfamily
You will be given a sentence and your task is to determine the meaning of the target word \{word\} from the following options: meaning 1 - \{meaning\_1\}; meaning 2 - \{meaning\_2\}; ... meaning n - \{meaning\_n\}. 
You need to determine the meaning of the target word \{word\} in the given sentence by providing the numerical label, without any further explanations. 
You should not provide any sort of explanation of your thinking process, and your final response should strictly follow this format: `Answer: [[label]]', where the label in the double brackets is the numerical label from the provided list, e.g., `Answer: [[1]]'. 

Target Word: `\{word\}' 

Sentence: \{sentence\}
\end{minipage}
\\

\bottomrule
\end{tabular}
\caption{Prompt templates used for the coarse-grained WSD experiments. The model is given a sentence, a target word, and a predefined list of candidate meanings. CoT+ prompts the model to explain which candidate meaning best fits the target word before giving the final numerical label, while CoT- requests only the final label.}
\label{tab:prompt-coarse-wsd}
\end{table*}

\end{document}